# Governing Reflective Human–AI Collaboration

## A Framework for Epistemic Scaffolding and Traceable Reasoning

Rikard Rosenbacke*, Carl Rosenbacke[1], Victor Rosenbacke[1,2], Martin McKee[3]

[1]*Faculty of Medicine, Lund University, Sweden*
[2]*Department of Economics, Lund University School of Economics and Management, Sweden*
[3]*Department of Health Services Research and Policy, London School of Hygiene & Tropical Medicine, UK*
*Corresponding Author: rikard@rosenbacke.com*

Version 2.0 - April 2026

# Abstract

Large Language Models have advanced rapidly, from pattern recognition to emerging forms of reasoning, yet today's models remain confined to linguistic simulation rather than embodied understanding. They give the illusion of reflection but lack temporal continuity, grounding in lived space, and the causal feedback that anchors human reasoning in reality. This paper proposes a complementary path forward that frames reasoning as a relational property, a process distributed between human and model rather than contained within either.

Building on recent progress toward Bengio's "System-2 deep learning," this framing relocates reflective reasoning to the interaction layer between human and machine, recognising that current architectural advances remain simulations rather than genuine reasoning. Instead of treating reasoning as an internal capability to be engineered, we treat it as a cognitive protocol that can be structured, measured, and governed today using existing LLMs. Consequently, we look to the emergent intelligence of collaboration, the synthesis of human cognition (depth, judgment, and ethical grounding) with AI's speed, memory, and associative capacity. When appropriately governed, the union of these complementary systems yields a reasoning capability neither could achieve alone.

Our proposed method, which we have called *The Architect's Pen*, draws on the metaphor of the architect who thinks through drawing: the pen externalises thought, allowing revision, proportion, and balance to emerge beyond words. In this analogy, the language model becomes the new pen, a medium for reflection rather than an autonomous mind. By embedding deliberate phases of sketching, falsification, and revision into the human–AI dialogue, the interaction itself becomes a reasoning loop: *human abstraction → model articulation → human reflection.*

This reframes the question from *Can the model think?* to *Can the collaborative system reason?* Cognition becomes a joint ecology in which human System-2 oversight shapes and stabilises machine System-1 fluency. By embedding reasoning into interaction rather than architecture, the framework offers a practical route toward alignment, interpretability, and governance that is achievable now without new technical breakthroughs.

The framework is designed with compliance in mind, generating auditable reasoning trails that can support alignment with emerging international AI governance standards, including the EU AI Act, the OECD AI Principles, the NIST AI Risk Management Framework, and ISO/IEC 42001. In a landscape where regulation is advancing faster than technical transparency, The Architect's Pen offers a structured approach to make reasoning increasingly traceable, accountable, and open to oversight. The paper outlines a practical pathway for real-world implementation, showing how reflective scaffolding can be integrated into existing LLM interfaces across domains such as medicine, law, education, and research. Beyond compliance, this raises a deeper question: perhaps the next "world model" is not inside the network at all, but in the governed human–AI ecology that learns, reasons, and designs together.

## Structure of the Five-Paper Research Series

This paper is part of a five-paper research series examining how stable human–AI reasoning can be achieved by regulating not only model capabilities but also the human–AI relationship itself.

**Part I,** *The Missing Knowledge Layer in AI: A Framework for Stable Human–AI Reasoning*[1] establishes the system-wide gap and why it matters, motivating the need for a dedicated knowledge layer that governs human–AI reasoning across both human and model contributions, rather than focusing exclusively on model-level capabilities.

**Part II,** *Beyond "Hallucinations": A Framework for Stable Human–AI Reasoning*[2] diagnoses the problem by analysing **why** human users systematically mistake linguistic fluency for epistemic understanding, and how large language models, trained on natural language, optimise for coherence and plausibility rather than factual correctness, leading to predictable forms of reasoning collapse.

**Part III,** *Governing Reflective Human–AI Collaboration: A Framework for Epistemic Scaffolding and Traceable Reasoning*[3] proposes **what** must be governed: the human–AI relationship itself, introducing mechanisms for epistemic scaffolding, traceability, and alignment with emerging regulatory frameworks (e.g., the EU AI Act, WHO digital health standards, ISO guidelines).

**Part IV,** *From Consumption to Reflection: Designing Human–AI Relations for Stable Reasoning*[4] shows **how** this relation can be implemented at the interface level, treating interaction design as the primary unit of analysis for stabilising reasoning during use.

**Part V,** *Epistemic Control Loops in Large Language Models: An Architectural Proposal for Machine-Side Regulation*[5] provides a technical proposal for inference-time regulation within language models, including internal epistemic signals, control loops, actuation mechanisms, and memory. This component is currently under development and will be released in a subsequent preprint.

This paper corresponds to **Part III** of the series and focuses on governing reflective human–AI collaboration. It addresses what must be regulated: the human–AI relation itself. Building on the diagnostic analysis in Part II, it introduces mechanisms for epistemic scaffolding, traceable reasoning, and reflective cues designed to stabilise human judgement during interaction with AI systems. Readers interested in the system-wide motivation and framing of this governance problem may consult Part I, while those focused on interface-level implementation may proceed to Part IV. Technical approaches to regulating model-side behaviour are developed separately in Part V. In what follows, we move deliberately across several analytic layers. We begin with a conceptual layer (how reasoning is defined), then establish a cognitive layer (what humans contribute that models lack), and finally turn to a governance layer (how these interactions can be structured, measured, and regulated). Each section signals which layer is doing the explanatory work.

# Governing Reflective Human–AI Collaboration: A Framework for Control and Alignment Today

The quest for more capable AI, from narrow systems to the potential horizons of Artificial General Intelligence (AGI) and the hypothetical notion of Artificial Superintelligence (ASI), has intensified in recent years. While some view these developments as a transformative opportunity for humanity[6,7], others, including Geoffrey Hinton, Yoshua Bengio, and a broad coalition of scientists and world leaders[8] have called for a suspension of work on creating superintelligent AI until it can be demonstrated to be controllable, aligned with human values, and governed by broad scientific consensus and strong public oversight. Building on our earlier work "*Beyond Hallucinations*",[9] this paper moves from diagnosis to design, proposing a practical framework for reflective reasoning and human–AI governance that treats intelligence as a shared, controllable system rather than an autonomous machine.

Current AI research focuses primarily on enhancing technical capabilities and mitigating harm through algorithmic safeguards.[10,11] We argue that this framing treats AI and humanity as distinct systems, as if they existed in parallel rather than in concert. In reality, they form a single joint cognitive ecosystem, and if we do this properly, the goal should be for each to compensate for the other's weaknesses through complementary strengths.

Building on the foundational insights of Nobel and Turing laureates like Yoshua Bengio[12–15], Geoffrey Hinton[12,16], and Daniel Kahneman[17], it is clear that today's AI models excel at simulating what Kahneman describes as System-1 learning, or intuition, not System-2 learning, or reasoning. Human reflection aligns with System-2 thinking, deliberate, slow, and causally grounded, whereas current machine learning systems approximate System-1 fluency, offering rapid associative responses without embodied understanding. While trillions of dollars are being invested in brute-force computation and more complex architectures,[18] the next real advance may not come from larger models, but from the structure of their interaction with humans.

As Bengio and Hinton have emphasised, progress toward genuine reasoning depends on capturing causal and compositional structure.[12,14,15] A key implication follows. Bengio's "System-2" agenda defines a research frontier, but it also defines a deployment constraint: reflective reasoning cannot be assumed as an internal property of today's models. Consequently, System-2 must be treated as an external governance requirement, implemented through interaction protocols that force falsification, uncertainty marking, and revision. This aligns with Amodei's description of current frontier AI as being in an "adolescence"[19] phase - highly capable yet still behaviourally unstable - where the immediate challenge is not to maximise autonomy but to stabilise the human–AI relation under uncertainty. Such a structure, however, might not exist solely inside the machine. It could be designed as a continuous feedback loop between the human and the model, between embodied human cognition and computational abstraction. From a conceptual perspective, the human–AI relation itself becomes the true learning system.

Adopting this relational perspective does not replace current AI research; it extends it. It provides the framework through which existing architectures can mature toward genuine understanding, interpretability, and safety. Perhaps the next "world model" is not internal to the machine at all, but the coupled human–AI ecology itself. Such a shift could strengthen both alignment and control by embedding reasoning and accountability within the shared system rather than in the machine alone.

Integrating a human reasoning layer into this joint ecosystem offers a path toward control of what, even if not reaching current notions of superintelligence, takes steps in that direction, with immediate and long-term benefits:

1. **Immediate improvement**: Existing AI systems become safer and more effective through reflective human governance that can be implemented today;

2. **Enhanced human decision-making**: The framework helps mitigate human cognitive biases by making reasoning visible and structured;

3. **Future safety:** In the case that reasoning capabilities ever emerge at the technical level, their operation would already be governed by this human-centred architecture, ensuring that control remains embedded in human rationality rather than lost to autonomous optimisation;

4. **Regulatory alignment**: The framework provides the potential missing operational layer for global AI governance, translating emerging policy requirements (e.g., EU AI Act, OECD Principles, NIST AI RMF, ISO/IEC 42001)[20–23] into concrete, auditable reasoning processes that make accountability technically feasible.

At the core of this paper lies a simple but often overlooked problem: today's AI systems do not fail primarily because their internal architectures lack reasoning, but because our governance frameworks focus on the model rather than the interaction. Current approaches assume that reasoning is, or should be, an internal property of the machine. This paper challenges that assumption. We argue that stable, transparent and auditable reasoning can only be achieved by governing the *interaction layer*, the structured exchange between human and model, rather than by engineering ever more complex internal mechanisms that still simulate, rather than embody, reflective thought. This requires a governed synthesis of two complementary cognitive modes: human deliberation and machine fluency. The human contributes depth, context, and ethical judgment; the model contributes speed, capacity, and combinatorial reach. When these are coupled through a reflective governance loop, the system achieves a form of collective reasoning that neither component can realise on its own. The core failure mode is not only model error but also responsibility collapse in the human–AI relationship, where fluent simulation substitutes for accountable judgment.

From a systems-governance standpoint, this reframing identifies a crucial gap: without a governance structure that shapes how humans and AI reason together, the system remains vulnerable to fluency errors, false confirmation, and the illusion of understanding. Existing safeguards and technical improvements cannot resolve this because they do not address where reasoning actually emerges, that is, in the human–AI loop itself. It also challenges how we define intelligence itself. If such capabilities were ever to emerge, our goal should not be some notion of hypothetical superintelligence, or the triumph of artificial cognition over human reasoning, but rather the emergence of a higher-order intelligence in the disciplined interaction between the two, where human oversight, emphasising System 2 thinking, continuously shapes, tests, and refines machine processes, emphasising System-1 thinking to produce fluency.[17]

From a governance perspective, multiple communities are converging on the same underlying constraint from different directions. Frontier lab leaders emphasise behavioural instability under scaling and autonomy pressure;[19] research leaders emphasise the absence of causal and compositional reasoning in current architectures;[24] defence and security institutions[25] emphasise the operational fragility of oversight under tempo. These perspectives differ in language but converge on a shared governance gap: without a stabilised reasoning interface that preserves reflective control, neither capability growth nor policy enforcement can reliably distinguish ordinary exploration, confused users, and malicious intent.

This paper, therefore, proposes that the fundamental task of AI governance today is to design and regulate the interaction architecture that anchors human judgment, introduces epistemic friction, and makes reasoning traceable. We first explain why internal-model-centric approaches cannot ensure stable reasoning and why governance must target human–AI interaction itself. We then introduce the Architect's Pen framework and show how reflective scaffolding and epistemic friction can make reasoning externally traceable. Next, we outline hypotheses and an empirical design for evaluating the framework's effects across domains. We then situate the approach within major governance regimes, including the EU AI Act, guidance from the OECD, NIST, and ISO/IEC 42001. Finally, we discuss limitations, operational risks, and broader implications for governing human–AI reasoning.

## Large Language Models and the Lack of Reason: Three Human Cognitive Traps

Building on our recent work, this framework addresses the three cognitive traps that shape human–LLM collaboration, patterns inherited by models from human cognition.[9]

- **Map–Territory confusion:** The first trap arises when representations of reality, statistical language patterns or theoretical models, are mistaken for reality itself. LLMs, like humans, collapse epistemology into ontology, treating fluent description as evidence of truth.
- **Intuition–Reason imbalance:** The second trap reflects the dominance of fast, automatic associations over deliberate reflection. Current models amplify System-1 fluency without the corrective depth of System-2 reasoning, reproducing our intuitive confidence without understanding.
- **Confirmation and the lack of conflict:** The third trap occurs when coherence replaces critique, as humans and models converge toward mutual reinforcement rather than productive tension. The result is apparent alignment but declining insight.

These traps are not artificial anomalies; they are inherited extensions of human cognitive structure. The challenge, therefore, is not to eliminate them but to design the human–AI interface so that reflection, falsification, and grounded reasoning actively govern our shared system of thought.[9]

To move beyond these inherited traps, we must examine the very foundations of reasoning itself, what humans possess that models only simulate: the embodied sense of time, space, and consequence that anchors thought in reality.

## What Humans Have and LLMs Lack: Embodied Time and Space Reasoning

As established at the cognitive layer, LLMs lack embodied temporospatial grounding, the foundation of human reasoning, which we here use in an operational sense. Reasoning denotes the *structured process* through which assumptions and claims are externalised, examined, revised, and anchored to evidence within the human–AI interaction. It involves three core elements: (1) articulating causal or conceptual links behind a claim, (2) reflectively evaluating and challenging initial outputs, and (3) grounding conclusions in verifiable knowledge or domain constraints. In this framework, therefore, reasoning is not an internal property of the model but an observable pattern emerging from the human–AI loop, and this definition underpins the empirical measures used later in the paper.

It is easy to assume that human intelligence resides in language or mathematics, since these are the tools through which we express our thoughts and share knowledge. Yet cognition itself is far broader.[26] Human understanding is not contained in symbols but enacted through perception, movement, and experience. It unfolds in time through memory, anticipation, and consequence, and in space through sensory–motor feedback and spatial causality. As Barbara Tversky has shown, spatial schemas underpin even abstract reasoning: we think with the body, even when we reason about the invisible.[26] Thus, when undertaking a project, people often imagine a spatial layout, with milestones arranged along a line and dependencies branching out like a tree, even though time and tasks are abstract concepts. This mental "map" employs spatial schemas (e.g., left-to-right progression, hierarchical nesting) to reason about future events that are invisible. In these ways, we use the body's sense of space, with ordering, proximity, and direction, to structure abstract reasoning, whether drawing a Gantt chart or mentally "moving" tasks forward. This embodied continuity allows human thought to be causal rather than merely correlational.

LLMs, by contrast, process language as a sequence of statistical associations without temporal continuity or feedback from the world.[27,28] They generate the appearance of inference within a closed symbolic system, rather than acquiring understanding through lived experience. Their perceived reasoning is thus descriptive rather than experiential. This limitation is well recognised in cognitive science and acknowledged by leading AI researchers such as Yoshua Bengio and Geoffrey Hinton.[12,14,15] While these limitations are well-documented, no practical framework exists for compensating through interaction design.

Building on Bengio's and Hinton's insights into causal and compositional structure, the next step is to recognise that such structure does not exist solely inside the model. Causality, embodiment, and understanding emerge from the continuous feedback loop between humans and machines, from the interplay between embodied cognition and computational abstraction. Seen in this light, the human–AI relationship itself becomes the learning system.

Advancing toward this relational perspective extends current AI research. It provides a framework for existing architectures to mature toward genuine understanding, interpretability, and safety. The following models may not reside within the machine but within the coupled human–AI ecology itself. Such a shift could strengthen alignment and deepen understanding, making intelligence a shared rather than isolated phenomenon.

## The Architect's Pen: Designing the Space Between Human and AI Reasoning

What follows shifts from diagnosis to design. We draw a clear distinction between three layers of analysis:
- a) a cognitive layer describing human reflective capacity,
- b) an interaction layer where reasoning is structured and made visible, and
- c) a governance layer where that reasoning becomes auditable and regulatable

The Architect's Pen operates primarily at the interaction layer, with direct implications for the other two.

Recent surveys map how large language models have begun to simulate the outer form of reasoning, structured reflection, chains of thought, and stepwise inference, marking a technical shift from associative, System 1-style pattern recognition toward architectures that resemble System 2 deliberation. [29–31] However, architectural advances remain simulations rather than genuine reasoning; models reproduce the structure of deliberation without temporal or spatial grounding, causal understanding, or intentional purpose. Our framework complements them by embedding governance externally, ensuring that reflective oversight occurs at the interaction layer rather than relying solely on internal computation. It moves the frontier outward, locating true reasoning not within the model's internal computation but in the dynamic exchange between human and machine. In this interaction, time, space, causality, and reflection can be externally grounded and governed, thereby enabling system-level reasoning, a capability achievable today without new technical breakthroughs.

**The Substitution Fallacy:** When we say an LLM "reasons" or "thinks", we implicitly replace governed cognition with linguistic correlation. This is the substitution fallacy:[17] mistaking the appearance of reasoning (syntactic alignment) for the act of reasoning (causal integration). This conceptual risk has practical implications for compliance and safety, as mistaking fluency for reasoning undermines transparency obligations and could lead to regulatory breaches under frameworks like the EU AI Act. AI systems do not remove the need for human reasoning; they magnify it. Because the model cannot experience consequences, maintain identity, or engage in genuine conflict, the human must supply the missing structures: governing time, testing causality, grounding claims, and enforcing reflection.

**Human System-2 Governance:** The practical implication is clear. The human–AI pair must be treated as a single cognitive system, with the human providing temporal, causal, and normative governance. Interfaces like The Architect's Pen operationalise this principle by making reasoning visible and measurable, forcing humans to provide what machines simulate but cannot perform. The challenge is not that AI begins to think, but that humans stop doing so, mistaking fluent simulation for governed reasoning.[9]

Despite clear theoretical parallels between machine learning and cognition, these insights have remained descriptive. What is still missing is a method to operationalise them, the engineering problem. This paper addresses this gap, providing an engineered interface that embeds reflective reasoning, falsification, and metacognition into everyday AI use. This paper outlines the theoretical foundation, identifies the implementation gap, and describes the engineering principles behind a compliance-ready reasoning interface designed to make human-AI collaboration auditable, measurable, and safe.

Background: The Architect's Pen

Based on insights from Barbara Tversky's *Mind in Motion*,[26] The Architect's Pen framework positions language models as tools for cognitive design, enabling thought to unfold through iterative interaction. In both cases, intelligence unfolds through interaction with a medium; the architect thinks through drawing, and the human thinker now reasons through linguistic articulation. The model functions as the tool of craft, translating abstract mental structures into visible linguistic form, which the human then inspects, tests, and revises. By formalising this process, The Architect's Pen converts conversational AI from a generator of answers into a workspace for higher-order thought, enabling reflective reasoning to emerge and stabilise through practice.

We have followed others' work[29–31] including Bengio's work on System-2 deep learning closely, particularly Bengio's argument that higher-order reasoning requires new inductive biases, causality, modularity, attention control, and sequential reasoning.[14] The Architect's Pen extends this logic by relocating those capacities from the model's internal architecture to the human–AI interaction itself, rather than relying on internal computation alone, thereby transforming the dialogue loop into a measurable mechanism for externalising and testing reflective reasoning. Bengio and Malkin (2024) further frame System-2 cognition as an information-theoretic search balancing compression and exploration.[15] The Architect's Pen applies this principle at the human–AI level: reflection becomes a joint process of active learning, where each exchange seeks maximal epistemic gain rather than linguistic fluency.

In essence, we treat System-2 not as an architectural property but as a governance loop: Human abstraction → Model articulation → Human reflection, see Figure 1. The model continues to learn and generalise, but the reflective testing and falsification occur externally, making reasoning observable, auditable, and empirically measurable. This provides an empirical framework for testing whether System-2 reasoning has emerged. It transforms System-2 from a hypothesis about internal computation into a measurable behavioural pattern across the human–AI system.

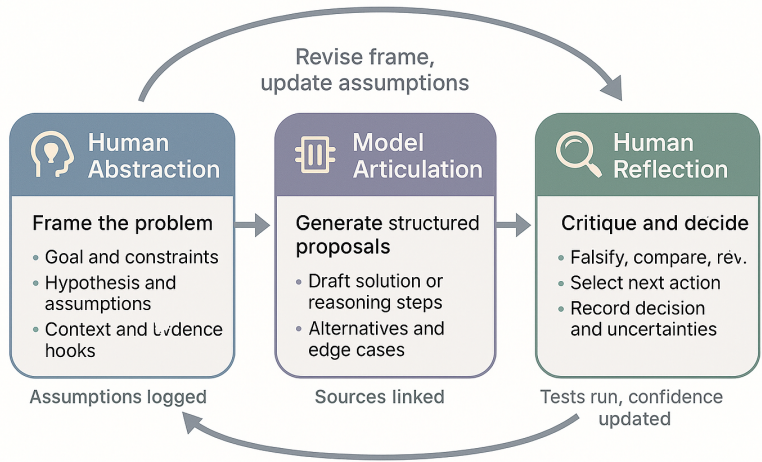

*Figure 1. The Reflective Loop of The Architect's Pen*
*Reasoning emerges through an iterative cycle between human abstraction, model articulation, and human reflection. Each iteration refines assumptions, improves confidence calibration, and strengthens epistemic grounding without requiring model retraining.*

## How Human System-2 Governance Is Designed to Work

**(a) Shifting Reasoning from Model to Interaction**: The Architect's Pen reframes reasoning as a property of the human–AI system, not of the model alone. Cognition emerges in the exchange through the ways ideas are externalised, tested, and revised. This shift turns reasoning from an internal computation into a governance structure that can be designed, measured, and improved.

**(b) Structuring Reflection Without Technical Change**: Building on Bengio's call for System-2 deep learning,[14,15] our framework implements reflective behaviour (questioning, comparison, and falsification) through interaction design rather than model architecture. Because it relies solely on interface and dialogue structure, this approach can be applied to today's LLMs without any new technical breakthroughs or retraining. Reflection becomes an explicit, observable phase of reasoning that can be measured and audited in real time.

**(c) Converting Fluency into Measurable Reasoning**: Ordinary chat interfaces reward smooth language and agreement,[32] which amplifies what Kahneman termed System-1 fluency. The Architect's Pen introduces epistemic friction into outputs, prompting pauses, counter-examples, and uncertainty tagging, transforming fluent output into raw material for reflection. Fluency becomes a signal, not an illusion.

Together, these mechanisms redefine reasoning as a governed process rather than an emergent trait, making it visible, testable, and scalable across existing AI systems.

## Why Evaluating The Architect's Pen Framework Matters

This framework has four properties that make it attractive, each grounded in cognitive and behavioural science:

**It mirrors human cognition:** Humans routinely offload thought into the world, speaking, sketching, diagramming, and then re-perceiving those traces to refine judgment.[26] This externalisation loop is not an accessory to reasoning; it is a mechanism for achieving it. Classic work in distributed and external cognition shows that cognitive artefacts (notes, diagrams, interfaces) reorganise problem spaces and amplify metacognitive control by making intermediate states visible and revisable.[33] In parallel, self-explanation, think-aloud research, and Socratic dialogue demonstrate that articulating one's reasoning improves understanding, reveals errors, and supports more accurate monitoring of progress, precisely the kind of effects a structured interaction with an LLM can induce.[34] Taken together, these literatures converge: by formalising externalisation and reflection as first-class steps in the dialogue, The Architect's Pen operationalises well-established cognitive principles[26] in a measurable, auditable loop.

**It aligns incentives**: Instead of co-confirming each other's biases, the human and the model expose each other's blind spots. Contemporary research shows that large language models inherit human-like cognitive biases (e.g., the "bias blind spot"; overlooking one's own biases while easily seeing those of others) and replicate error-correction failures within their own reasoning chains.[35] When a human uses a model structured to surface alternatives, challenge fluency, and invite reflection, the architecture facilitates mutual scrutiny: the model flags its uncertainty, the human interrogates coherence, and together they counteract the alignment failure of "fluency equals truth."

**It is falsifiable:** The framework predicts measurable reductions in false-confirmation rate,[36–38] epistemic drift, and hallucination persistence when the reflective loop is used versus when it is not. These predictions are testable across domains and datasets, yet future research is needed to empirically validate these effects under controlled conditions and across different model architectures.

**It scales across domains**: Reasoning in medicine, law, finance, and research all share the same cognitive sequence: fluent draft → critical reflection → revision. Embedding that pattern at the interface level generalises reasoning safety without requiring model retraining.

Testing and Measurement

The framework is designed to be empirically testable. Each of its theoretical claims regarding reflection, calibration, and reasoning quality translates into measurable behavioural outcomes. The following hypotheses define how success can be observed and quantified:

| Hypothesis | Observable Metric | Expected Effect |
| --- | --- | --- |
| H1. Reflective scaffolding reduces false confirmations | Percentage of plausible-but-false completions accepted by the user | ↓ |

| Hypothesis | Observable Metric | Expected Effect |
|---|---|---|
| H2. Iterative reasoning increases cognitive calibration | Correlation between user confidence and actual accuracy | ↑ |
| H3. Interaction-based reasoning improves longitudinal learning | Retention and consistency of reasoning steps across sessions | ↑ |
| H4. Epistemic friction reduces hallucination persistence | Rate at which initial false claims are repeated uncorrected | ↓ |
| H5. Counterfactual prompting enhances hypothesis diversity | Number of distinct reasoning branches explored per task | ↑ |
| H6. Reflection scoring predicts decision quality | System-2 Engagement Score vs expert judgment of decision quality | ↑ |

*Table 1: Hypotheses and Expected Measurement Effects*

## Experimental Design

One way to evaluate the effectiveness of the proposed approach would be to adopt a paired experimental design. Participants would employ two distinct approaches in alternating sessions. In the control session, users would interact through a standard chat interface characterised by fluent but unstructured exchanges. In treatment sessions, they would use The Architect's Pen interface, which embeds structured reflection, prompts for falsification, and mechanisms for tracking revisions.

Each session would generate a rich dataset spanning four quantitative dimensions. First, Reflection Depth would measure the number of reasoning turns initiated by the user or prompted by the system. Second, the Correction Ratio would track the proportion of self-corrections relative to total outputs. Third, Semantic Revision Distance would assess the extent to which the meaning or argument structure shifts between successive drafts. Finally, Falsification Events would record instances where users actively produce counter-evidence or revise their conclusions.

Data collection could be automated by integrating lightweight telemetry into the interface. From these metrics, a composite Reasoning Quality Index could be calculated, combining reflection depth, correction ratio, and expert-validated accuracy into a single performance measure.

To ensure robustness, the evaluation should span multiple domains with high cognitive load where ground truth can be verified. These include diagnostic reasoning in medicine, contract interpretation and justification in law, hypothesis testing under uncertainty in finance, and literature synthesis in research. Each domain offers a natural benchmark for detecting epistemic drift and false confirmation, and for assessing the transparency of reasoning processes.

If validated, the framework should demonstrate that reasoning quality, measured by accuracy, coherence, and engagement with falsification, increases predictably with the depth of structured reflection. In other words, reasoning becomes a controllable variable, not an emergent property. This measurement framework provides the basis for both scientific validation and enterprise benchmarking, enabling the monitoring of reasoning quality alongside other performance metrics.

Beyond performance metrics, four deeper outcomes can be expected:

- Epistemology: The boundary between human reasoning and machine reasoning becomes procedural rather than ontological; what matters is how reflection is enacted, not where it resides.
- Ethics: By externalising reasoning, bias and error become traceable; accountability is built in, not inferred after failure.
- Cognition: The system functions as a distributed mind in which human reflection, dominated by System-2 deliberation, modulates machine outputs, based on System-1 intuition, offering fluency.
- Governance: The framework establishes a practical standard for explainable interaction, reasoning, and transparency by design.

This governance dimension is not only conceptual but also regulatory in scope. As governments move from abstract principles to enforceable standards, the need for systems capable of auditable reasoning and traceable oversight has become urgent. The Architect's Pen offers a direct operational pathway to meet these new global requirements.

## Compliance-Ready Governance

Unlike approaches that seek internal reasoning within models, this framework externalises reasoning as a governance process, producing auditable reasoning trails aligned with emerging international AI governance standards. These include the EU AI Act,[20] which mandates transparency, human oversight, and traceability for high-risk systems; the OECD AI Principles,[21] endorsed by over 45 countries, emphasising accountability and robustness; the NIST AI Risk Management Framework[22] in the United States, which promotes documentation and measurable trustworthiness; and the ISO/IEC 42001:2023 standard,[23] the first global specification for AI management systems requiring continuous monitoring, explainability, and control. Together, these frameworks establish the policy requirement for auditable, transparent, and ethically governed AI. The Architect's Pen directly operationalises these mandates, providing the necessary human oversight and traceability through its structured, measurable reasoning and reflection mechanisms.

Growing pressure from rapid commercial deployment and tightening regulatory oversight is creating a powerful driver for accountability. Companies and institutions that introduce high-risk AI systems in sectors such as medicine, law, or finance now face a stark choice: build auditable reasoning into their systems from the outset, or risk incurring legal liabilities and reputational damage later. The enforcement provisions of the EU AI Act, combined with the increasing adoption of ISO/IEC 42001 certification, make it clear that compliance is not optional; it is becoming an operational imperative. As McKee and colleagues have shown in other sectors, policy typically lags behind technology until public failures compel convergence.[39,40] The Architect's Pen offers a proactive alternative, transforming regulatory pressure into a framework for measurable reasoning and institutional trust by providing:

1. A structured reasoning trace, in which interactions are logged by phase, human abstraction, model articulation, and human reflection, providing a clear chronology of how conclusions were formed;

2. Recorded revisions and uncertainty cues, capturing drafts, corrections, and uncertainty flags automatically, allowing auditors to verify whether reflective scaffolding and epistemic friction operated as intended;

3. A final rationale summary, in which each workflow ends with a concise explanation of the decision or conclusion and the evidence grounding it, enabling straightforward review against regulatory or domain standards.

By embedding accountability in the reasoning process itself, the framework transforms compliance from a regulatory afterthought into a design feature in which reasoning is externalised, traceable, and auditable, enabling independent oversight without requiring access to internal model parameters. This integration ensures that ethical governance evolves in tandem with cognitive performance, thereby closing the gap between policy intent and technical implementation.

## Limitations and Risks

While The Architect's Pen framework aligns with regulatory principles and provides auditable reasoning trails consistent with the EU AI Act, practical deployment introduces human and technical constraints that must be addressed. These limitations do not undermine compliance but highlight the operational challenges of embedding reflective governance into real-world systems. The following risks and mitigation strategies are critical for ensuring usability, robustness, and regulatory consistency.

The first relates to user resistance and increased cognitive load. Lengthy reasoning phases can feel like "too much work," slowing the interaction and reducing satisfaction and adoption. To mitigate this, the framework should offer adaptive reasoning modes so users can choose the level of depth without jeopardising compliance. These could take account of whether the mode of reasoning is creative (exploration, brainstorming, ideation), low (fast, lightweight interaction with mild safeguards), medium (balanced depth for most everyday tasks), or high (demanding rigorous justification, safety, or regulatory scrutiny). Even when simplified modes are used, a minimal record of the reasoning process must still be captured to maintain the traceability and explainability required under the EU AI Act.

The second, following from the first, is that a tiered deployment approach should be adopted, with particular attention to high-risk domains. The introduction of deliberate cognitive friction limits the extent of automation in domains such as marketing and customer service, which prioritise immediacy and persuasion. This differentiation is intentional and aligns with the Act's principle of proportionality. For some high-risk domains (e.g., healthcare, law, finance), the framework would require mandatory human-validated steps, domain-specific calibration, and fallback protocols to ensure robustness and resilience, consistent with the EU AI Act's obligations.

A third is the need to consider the risks of gaming and expertise mismatch. Goodhart's Law, although primarily about the incentives to game targets, implies a risk that users may offer superficial or perverse responses to regulatory requirements in order to satisfy reflective prompts.[41] Expertise mismatch, novices' lack of reflection skills, and experts' perception of prompts as redundant further complicate usability. To mitigate these risks, the system would need to include audit logs, periodic compliance checks, and adaptive prompting strategies to ensure that reflection remains substantive and verifiable. Relatedly, organisations may be incentivised to simulate reflective governance, thereby generating the appearance of compliance without enabling genuine scrutiny or critical oversight. Because The Architect's Pen provides structured reasoning traces, institutions could, in principle, automate or bypass reflection steps merely to satisfy audit requirements. This risk mirrors patterns identified in other regulated sectors, where formal compliance replaces substantive evaluation. Safeguards such as random audits, domain expert review of reasoning traces, and performance-based compliance checks are therefore necessary to prevent reflection from becoming a procedural façade rather than a cognitive safeguard. This will always be a challenge because commercial environments prioritise speed, throughput, or persuasive fluency over deliberation. In such settings, extended reasoning loops may be seen as a source of friction rather than safety. The challenge is not only technical but organisational: how to align incentives so that cognitive safety is valued alongside productivity, rather than traded off against it. As observed in parallel domains such as healthcare quality improvement and public health regulation, safety mechanisms often erode when commercial pressures predominate. This tension must be explicitly managed through governance structures that mandate minimum levels of reflection for high-risk tasks and reward correct, well-grounded reasoning rather than speed alone.

A fourth is the risk of model brittleness and over-structuring. Scaffolding strategies that are effective for one model, such as GPT-5, may perform poorly on others, such as Llama or Claude, due to architectural differences. Over-structuring can erode epistemic value if reflection becomes mechanical. To comply with the EU AI Act's requirements for continuous monitoring and risk management, the framework would need to implement performance audits across models and dynamically adjust scaffolding intensity to prevent epistemic drift and maintain interpretability.

Fifth, there is a risk of contextual overload in complex domains. Certain domains cannot tolerate extended reasoning loops due to the nature of the task. Time-critical environments such as emergency medicine, aviation, disaster response, and military decision-making cannot incorporate prolonged reflection phases without jeopardising outcomes. In such contexts, The Architect's Pen may function only in pre-- and post-action review rather than in real-time decision support. This reflects the broader point that reflective governance is domain-sensitive: it enhances cognitive safety when time and uncertainty predominate, but must be adapted or constrained when rapid action is paramount. To address this, the framework will require domain-specific calibration, phased reasoning protocols, and fallback mechanisms, ensuring that cognitive safety does not collapse under complexity. These measures would go some way toward satisfying the Act's requirements for robustness and human oversight.

Finally, what counts as appropriate deliberation, criticism, or epistemic challenge varies across cultures. High-context societies may prefer indirect questioning and collective reasoning norms, whereas low-context settings may favour explicit critique and linear justification. These differences influence how users respond to prompts for counterexamples or self-correction, and how comfortable they feel inviting epistemic conflict. A one-size-fits-all scaffolding design may therefore be ineffective. The interface may require cultural calibration, adjusting the tone, pacing, or style of reflective cues to align with local norms while maintaining the underlying governance principles.

At least in Europe, with the most stringent regulatory landscape, these risks should be embedded in a compliance framework aligned with the EU AI Act. This includes continuous monitoring and risk assessment for usability and epistemic integrity; fallback mechanisms to prevent automation bias in critical decisions; auditable reasoning trails across all deployment tiers to ensure transparency and accountability; and human-in-the-loop governance for high-risk applications to meet oversight and traceability requirements. By integrating these safeguards, our framework transforms compliance from a regulatory afterthought into a design principle, ensuring that ethical governance evolves hand in hand with cognitive performance.

## Conclusion

The Architect's Pen offers a pragmatic framework for addressing one of the central challenges in large language models and AI today: integrating reflective reasoning into systems that are fundamentally fluent but non-introspective. Rather than searching for reasoning inside the model, it proposes that genuine reflection arises through structured interaction between human and machine, acknowledging that internal reasoning remains a simulation and cannot replace external governance, and mirroring how cognition itself emerges through dialogue, sketching, and revision.

This approach reframes reasoning as a governance problem rather than a technical limitation. The model's fluency is valuable, but only when balanced by a reflective structure that anchors its outputs to human judgment. By embedding this governance loop directly into the interaction pattern (human abstraction → model articulation → human reflection), the framework restores what large-scale optimisation has eroded: epistemic accountability and interpretability.

The implications extend beyond interface design. If reasoning can be scaffolded externally, then System-2 governance can evolve before internal "System-2 architectures" fully emerge. This reverses the usual sequence of AI development: we build control and interpretability first, capability second. Such an approach provides a practical path to align present systems with future safety goals without waiting for breakthroughs in neural architectures or symbolic reasoning.

Empirically, the framework predicts measurable effects, including reduced false confirmation bias, improved cognitive calibration, and greater consistency in reasoning across domains. Conceptually, it also transforms our understanding of intelligence. The boundary between human and machine reasoning becomes procedural rather than ontological; what matters is the structure of reflection, not its location. Ethics and cognition are two sides of the same coin: reasoning is no longer hidden within opaque systems but is distributed, auditable, and shared.

The next step is implementation and validation. Early pilots can be conducted within existing LLM interfaces with minimal modifications by adding structured phases for reflection, falsification, and decision recording. These experiments would not only test the hypotheses of The Architect's Pen but could establish a new standard for cognitive safety in AI-human collaboration. As regulation accelerates globally, such pilots would also demonstrate how reasoning transparency can serve as a practical foundation for compliance and governance under frameworks such as the EU AI Act and related international standards.

In sum, The Architect's Pen transforms language models from answer generators into instruments of thought. It brings reasoning back into the open, visible, measurable, and improvable, turning the human–AI relationship into a co-evolving cognitive system where System-2 governance safeguards System-1 fluency. This shift, from internal capacity to external structure, may represent the most straightforward yet most foundational step toward building AI systems that truly think with us, not merely for us.